\documentclass{article}
\usepackage[T1]{fontenc}
\usepackage{graphicx}
\usepackage{url}

\def\toprule{}
\def\midrule{\hline}
\def\bottomrule{}

\begin{document}

\title{Ontological Relations from Word Embeddings}

\author{Mathieu d'Aquin and Emmanuel Nauer\\
K Team, LORIA, Université de Lorraine\\
Nancy, France}
\date{August, 2024}

%\author{First Author\inst{1}\orcidID{0000-1111-2222-3333} \and
%Second Author\inst{2,3}\orcidID{1111-2222-3333-4444} \and
%Third Author\inst{3}\orcidID{2222--3333-4444-5555}}

%\institute{Princeton University, Princeton NJ 08544, USA \and
%Springer Heidelberg, Tiergartenstr. 17, 69121 Heidelberg, Germany
\maketitle              % typeset the header of the contribution
\begin{abstract}
It has been reliably shown that the similarity of word embeddings obtained from popular neural models such as BERT approximates effectively a form of semantic similarity of the meaning of those words. It is therefore natural to wonder if those embeddings contain enough information to be able to connect those meanings through ontological relationships such as the one of subsumption. If so, large knowledge models could be built that are capable of semantically relating terms based on the information encapsulated in word embeddings produced by pre-trained models, with implications not only for ontologies (ontology matching, ontology evolution, etc.) but also on the ability to integrate ontological knowledge in neural models. In this paper, we test how embeddings produced by several pre-trained models can be used to predict relations existing between classes and properties of popular upper-level and general ontologies. We show that even a simple feed-forward architecture on top of those embeddings can achieve promising accuracies, with varying generalisation abilities depending on the input data. To achieve that, we produce a dataset that can be used to further enhance those models, opening new possibilities for applications integrating knowledge from web ontologies. 
%\keywords{Ontologies \and Language Models \and Relation Learning.}
\end{abstract}

\section{Introduction}

Word embeddings~\cite{goldberg_neural_2017} are vector representations of words in a text, used, in particular, to perform learning tasks in natural language processing. Many of the embeddings used today are produced by neural (large) language models such as BERT~\cite{devlin_bert_2019}. These are neural network models trained on particular tasks (such as masked-language modelling, i.e. predicting masked tokens in a given text) and from which selected hidden layers can be used as embeddings. Interesting properties have been demonstrated for embeddings produced by different models, in particular, in representing the meaning of the words within the embedding space. For example, words that are semantically similar have been shown to generally have similar embeddings~\cite{mikolov_distributed_2013}. The main question that is asked in this paper is whether they encapsulate enough information to cover relations beyond semantic similarity. We particularly want to check whether embeddings from various models could be used to derive ontological relations between entities represented by their labels, such as the subsumption relation (subclass, subproperty), equivalence, or the ones connecting properties to their domains and ranges. 

To achieve that, we test whether simple models taking as input the average embeddings of the names and comments of classes and properties from popular upper-level and general ontologies are capable of predicting the relations (direct or inferred) existing between those entities. We constitute a dataset of relations (materialised by inference at the level of RDF entailment rules~\cite{w3c_rdf_nodate}) from five of those ontologies, having computed the embedding vectors for each entity in each ontology using four different (transformer-based) models. 

We obtain promising results with respect to the accuracy of the prediction, demonstrating the potential for knowledge models capable of encapsulating and extrapolating ontological relations in large open domains. We also study the generalising ability of each model by cross-validating them against each other's ontologies and report on the performance of a combined model, providing a first baseline for future work to build upon, using other larger language models, different architectures, and integrating more ontologies in their training set.

% The paper is organised as follows... 

\section{Related work}

A lot of attention has been given recently to the whole area of knowledge graph embeddings (see, for example,~\cite{paulheim_embedding_2023}) in particular for their use in link prediction tasks (see, for example,~\cite{zha_inductive_2022}). Knowledge graph embeddings are methods (often based on neural models) to project the graph representation available in knowledge graphs onto a vector space in a way that should align with the structure and meaning of the graph. They are often used for link prediction, i.e. the task of predicting which entity might be related to which other in a graph since, by encapsulating patterns in the existing graph, they should be able to discover where missing relations might exist. These two categories of work are closely related to the work presented in this paper since they relate to the prediction of relations between entities in knowledge structures through the use of embeddings. Here, however, we focus on ontological relations as predictable from word embeddings applied to the textual representation of the entities. In other words, while knowledge graph embeddings, in a sense, distill the content of a graph to find ways to complete it, we aim to exploit knowledge already captured by pre-trained language models through word embeddings to predict ontological relations between classes and properties of web ontologies. 

Following the work on knowledge graph embeddings, ontology embeddings have been proposed that focus on the OWL language~\cite{chen_owl2vec_2021} or on particular description logics~\cite{kulmanov_embeddings_2019}. While ontological relations are considered, those works aim to create representations similar to knowledge graph embeddings but that are capable of capturing the semantics of higher-level formalisms used for ontologies. They therefore also accomplish a different task from the one endeavoured here, even if it is strongly related and possibly complementary. 

Much closer to our work are a few proposals for predicting ontological relations between entities using language models, generally or in specific domains (such as biomedicine~\cite{liu_concept_2020}). Much of these works tend to focus on the subsumption relation and on placing classes in the hierarchy of an ontology (see, for example,~\cite{ristoski_large-scale_2021}), while our aim is to predict a wider set of relations. In addition, they often rely on a so-called probing process to exploit the language model~\cite{he_language_2023}. This means that at least a part of the process is based on generating a prompt to a large language model and analysing the generated textual output to identity relations between entities mentioned in the prompt. What we propose here is more straightforward (add prediction layers on top of word embeddings extracted from the language models) and less dependent on necessarily noisy prompt generation and result interpretation mechanisms. Other approaches (such as~\cite{ushio_relbert_2023}) go through the step of fine-tuning the language model used for the task of predicting subsumption (sometimes based on a number of domain-specific ontologies~\cite{liu_concept_2020}), introducing a high demand for computational resources. 

\section{Overview}
\label{sec:overview}

\begin{figure}[t]
    \centering
    \includegraphics[width=\linewidth]{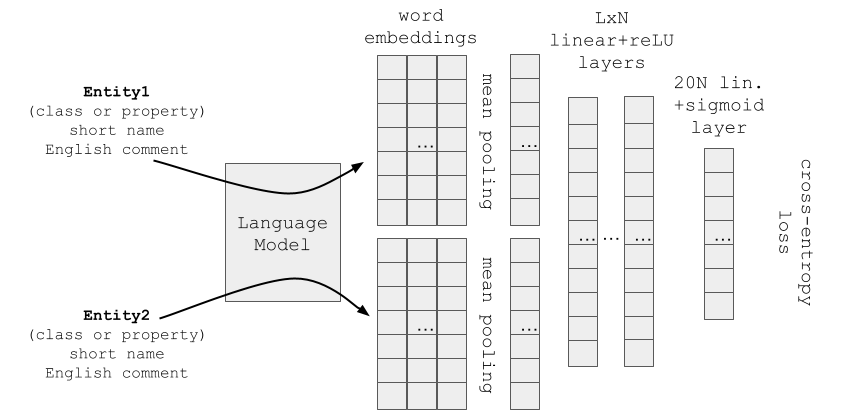}
    \caption{Overview of the architecture of the model predicting ontological relations (among 20) for two entities from the word embeddings of their short names and comments.}
    \label{fig:overview}
\end{figure}

Figure~\ref{fig:overview} provides an overview of the approach taken to predict ontological relations that might exist between two entities, represented as texts by their short name (the last part of their IRI) and comments in English (if they have one). 
The texts of the short name and of the comment of each entity are first run through a language model (in Section~\ref{sec:models}, we describe the four we tested) to obtain word embedding vectors for each of them, i.e. a word embedding is computed for each word of the short name and comment. To reduce those to one vector for each entity, we apply a mean-pooling operation, resulting in an average embedding for the entity, based on its textual representation. We then train a simple neural model made up of a few small fully connected layers with reLU activation functions (the number depends on the language model used and the input data, as discussed in Section~\ref{sec:training}). As there are 20 relations that can be predicted, which include directly stated relations (subclass, subproperty, equivalence, disjointness, domain, range), as well as their inverse relations (superclass, superproperty, etc.) and indirect or inferred versions of those relations, an output layer of size 20 is then constructed with a sigmoid activation function. It is compared for training, through a cross-entropy loss function, to the 20-sized binary vector which encodes the real relations existing in the input data. This architecture was chosen to be relatively simple and straightforward to setup, based on the idea that if interesting results can be obtained in predicting ontological relations with such an architecture, it will form a baseline for potentially more sophisticated models to improve upon. 

To carry out experiments on training such an architecture, labelled data need to be obtained. In the next section, we describe how we constructed datasets that associate average embeddings of the textual representations of pairs of entities with binary vectors representing their asserted and inferred relations in five different upper-level and general ontologies. We then discuss in Section~\ref{sec:results} the results of training, validating, and testing relation prediction models using different inputs (from the different ontologies and using the different language models), and evaluate those results using precision, recall, and F-score. 

\section{Dataset generation}
\label{sec:data}

In this section, we detail how datasets are created for the training and validation of relation prediction models. We start by describing the ontologies used and the process of extracting relations from those ontologies. We then briefly introduce the language models used to extract embeddings of the textual representation of entities, and finally what is included in the generated datasets. The generated datasets and intermediary structures computed to construct them are available on FigShare\footnote{\url{https://figshare.com/s/b216348f194ebad9d501}} and the code to generate them from an N-Triples file containing an ontology is available on github.\footnote{\url{https://anonymous.4open.science/r/LKM-B71B}}

\subsection{Ontologies}
\label{sec:ontologies}

As a basis for training and validation, we selected five ontologies. The reason for using those five is that, considering that the language models used were trained on open-domain texts, they should be better able to predict the relations existing in ontologies that are not specific to a particular domain. In other words, they rely, in the textual representation of the included entities, on a general vocabulary rather than a specialised one. We therefore used two upper-level ontologies and three general ontologies (i.e. domain-level ontologies but for the general domains). 
\begin{description}
    \item[DUL (upper-level):] The DOLCE+DnS Ultralite~\cite{presutti_dolce_2016} (DUL) ontology combines elements from the Descriptive Ontology for Linguistic and Cognitive Engineering (DOLCE) with the DnS (Descriptions and Situations) ontology patterns to provide a framework for interoperability between ontologies.
    \item[gUFO (upper-level):] gUFO\footnote{\url{https://nemo-ufes.github.io/gufo/}} is a lightweight implementation of the Unified Foundational Ontology (UFO) used for the development and integration of domain ontologies.
    \item[OpenVocab (general):] OpenVocab\footnote{\url{https://vocab.org/open/}} was an initiative to create a community-maintained general vocabulary. The editing interface has been closed, but an RDF version of the resulting vocabulary remains available.
    \item[Schema.org (general):] Founded by Google, Microsoft, Yahoo, and Yandex, schema.org~\cite{guha_schemaorg_2016} is a community-driven initiative to create schemas for structured data on the Web that can be used, in particular, by search engines to provide more precise and relevant results. 
    \item[DBpedia (general):] The DBpedia ontology\footnote{\url{https://www.dbpedia.org/resources/ontology/}} is the core ontology used to describe entities in the DBpedia knowledge graph. Originally created manually from common properties of Wikipedia infoboxes, it is now a regularly evolving community effort. 
\end{description}

These ontologies were downloaded from the most authoritative links that could be found for them in their latest version. A metadata file is included in the data repository for this paper indicating from which link they were obtained and at what date. The process described below to extract textual representations for and relations between the entities in those ontologies requires as input an RDF file in the N-Triples syntax. Some of those ontologies were, therefore, first converted into this format. 

\subsection{Extracting relations from ontologies}
\label{sec:relations}

The first part of the process of creating a dataset of relations between entities contained in the considered ontologies is to extract such relations, as well as those that can be derived from them. Some of the ontologies considered are relatively large (including a few thousand classes and properties), and relations between potentially every pair of entities had to be considered. To minimise the time to search the relations between two entities, each ontology is represented by a large $n{\times}n$ matrix where $n$ is the number of entities, and each cell of the matrix contains an integer where each bit represents one of the 20 possible relations we considered (e.g. a cell containing 0 means that there is no relation between the two entities). 

In order to include inferred relations as well as directly stated ones, we included in the process calls to an ad hoc forward chaining rule engine relying on rules to derive inverse relations from stated ones and to implement a relevant subset of RDF-S entitlements~\cite{w3c_rdf_nodate} (e.g. the transitivity of the \texttt{subClassOf} relation). Furthermore, as the matrix is created, each entity encountered is added to a table that associates with the IRI of the entity its index in the lines and columns of the relation matrix, as well as, if found, its English label and comment (based on the \texttt{rdfs:label} and \texttt{rdfs:comment} properties). Table~\ref{tab:relations} summarises the 20 relations considered and their frequency in the created datasets from each of the considered ontologies. Note, however, that the relations in DBpedia were randomly sampled to reduce their number and rebalance them, since some were significantly more represented than the others. Also, schema.org appears not to include domain and range relations, as those are made, in that ontology, with anonymous classes.

\begin{table}[h]
    \centering
    \caption{Relation counts in each of the ontologies.}
    {\small
    \label{tab:relations}
    \begin{tabular}{p{5mm}p{20mm}rrrrr}
\toprule
  \multicolumn{2}{l}{Property} 
& ~~DUL 
& ~~gUFO 
& ~~OpenVocab 
& ~~Schema.org 
& ~~DBpedia \\
\midrule
SbC: & subclass & 38 & 31 & 15 & 469 & 369 \\
SpC: & superclass & 38 & 31 & 15 & 469 & 369 \\
SbP: & subproperty & 51 & 7 & 13 & 68 & 424 \\
SpP: & superproperty & 51 & 7 & 13 & 68 & 423 \\
HD: & has domain & 68 & 25 & 55 & 0 & 67 \\
DO: & is domain of & 68 & 25 & 55 & 0 & 2 \\
HR: & has range & 71 & 24 & 48 & 0 & 131 \\
RO: & is range of & 71 & 24 & 48 & 0 & 40 \\
DW: & disjoint with & 9 & 15 & 7 & 0 & 8 \\
SA: & same as & 0 & 0 & 0 & 0 & 0 \\
E: & equivalent & 0 & 0 & 10 & 0 & 56 \\
ISbC: & inferred subclass & 158 & 88 & 29 & 1584 & 2360 \\
ISpC: & inferred superclass & 158 & 88 & 29 & 1584 & 2360 \\
IE: & inferred equivalent & 0 & 0 & 10 & 0 & 56 \\
ISbP: & inferred subproperty & 83 & 9 & 13 & 71 & 435 \\
ISpP: & inferred superproperty & 83 & 9 & 13 & 71 & 434 \\
IRO: & in the range of & 4305 & 532 & 67 & 0 & 2655 \\
IHR: & inferred has range & 172 & 70 & 51 & 0 & 271 \\
IDO: & in the domain of & 4536 & 591 & 75 & 0 & 1452 \\
IHD: & inferred has domain & 175 & 80 & 59 & 0 & 363 \\
\bottomrule
\end{tabular}
}
\end{table}

\subsection{Pre-trained Language Models for Word Embedding}
\label{sec:models}

In the process presented in Section~\ref{sec:overview}, we rely on a selection of popular language models to test which are more efficient in re-creating ontological relations. All those language models are pre-trained by their original authors and built on different architectures, although they are all neural models based on transformers, as summarised below. There are a number of other language models that could have been used and could be added in the future, but those represent a reasonable selection of what is openly available today. 

To extract embedding vectors for textual representations of classes and properties of ontologies with these models, we used the huggingface transformers library in Python\footnote{\url{https://huggingface.co/docs/transformers/en/index}}. In more detail, for every entity included in an index for an ontology, we first run the tokenized shortname of the entity through the model, obtaining an ``in context'' embedding vector for each of the words in the shortname by extracting the activations of the last layer before output (last hidden states). We reduce this set of vectors to one by averaging them (the mean-pooling step in Figure~\ref{fig:overview}). We apply the same process to the \texttt{rdfs:comment} of the entity if it has one, and average the name and comment vectors to obtain a final embedding vector of the textual representation of the entity. As a result, our process leads to a dataset including an embedding vector from each of the four language models for every class and property in each of the five ontologies considered here.  

\begin{description}
    \item[BERT:] BERT~\cite{devlin_bert_2019} (Bidirectional Encoder Representations from Transformers) was introduced by Google in 2018 and quickly became a reference language model for many tasks. It has the particularity of considering textual contexts in both directions (forward and backward). It was mostly trained on BooksCorpus and English Wikipedia, and produces embedding vectors of size 768, using the model identified as ``bert-base-uncased'' on huggingface.
    \item[RoBERTA:] RoBERTA~\cite{liu_roberta_2019} (Robustly Optimized BERT Pretraining Approach) is a proposal by Facebook AI to train the BERT architecture with different hyperparameters, obtaining better results, according to the paper, than the original BERT. RoBERTA also produces vectors of size 768 using the model identified as ``FacebookAI/roberta-base'' on huggingface. 
    \item[GPT2:] GPT-2~\cite{radford_language_nodate} is an advanced version of the original GPT (Generative Pre-trained Transformer) model produced by OpenAI. It is significantly larger than BERT (1.5B parameters, compared to 340M) and is trained on a more varied, larger, specially collected corpus (from web scraping). It also produces vectors of size 768 using the model identified as ``openai-community/gpt2'' on huggingface.
    \item[Llama2:] Llama2~\cite{touvron_llama_2023} is a set of open-source large language models of varying sizes (from 7B to 70B parameters) produced by Meta AI (formerly Facebook AI) intended to achieve performances comparable to large, popular close models (such as GPT3). It is trained on a large mix of publicly available data. Here, we use the 7B parameter version, producing embedding vectors of size 4096 using the model identified as ``meta-llama/Llama-2-7b-hf'' on huggingface. 
\end{description}

\subsection{Structure of the generated datasets}
\label{sec:datasets}

We created 20 training and validation datasets: one for each of the five ontologies together with one of the four language models considered. Each of these datasets includes the concatenation of the embedding vectors from the given language model of pairs of entities of the given ontology, associated with the binary vectors representing the presence or absence of ontological relations between those pairs. To separate the training set from the validation set, we first set aside 30\% or 40\% of the entities (depending on the ontology, to ensure that different types of relations are reasonably represented in the validation set) to create pairs for the validation set and use the remaining entities to form pairs for the training set. In the next section, we will, therefore, assess the performance of models having learnt from a training set, based on measuring precision, recall, and F-score on the corresponding validation set. Note that we only include in the datasets pairs of entities between which there exists at least one relation. Also, as mentioned in Section~\ref{sec:relations}, some relations in the datasets for the DBpedia ontology were randomly removed to reduce the imbalance between the different types of relation in that ontology. 

\section{Results}
\label{sec:results}

In this section, we present the results of training a number of models on the task of predicting the relations existing between pairs of entities, represented by the text of their names and, optionally, their descriptions as present in the \texttt{rdfs:comment} attribute. We first briefly describe the training process, show the results in terms of precision, recall, and F-score for the 20 combinations of ontologies and language models, and discuss those results. We then also show the results of cross-validating models trained on each of the ontologies against the validation set of the other ontologies on the best embeddings obtained in the previous step (Llama2). The goal here is to obtain results that enable us to gain an understanding of the generalisation capabilities of the models created. Finally, we also show the results of training one model from a combination of all the training sets from the five ontologies, to assess whether including relations from a larger, more varied set could lead to a globally more effective model not only of one ontology but of knowledge on the Web generally. 

\subsection{Training}
\label{sec:training}

As shown in Figure~\ref{fig:overview}, each model adds to the concatenation of the embedding vectors produced for a pair of entities a few (one to three) fully connected hidden layers with reLU activation and an output layer (of size 20) with sigmoid activation. The decision on the number of hidden layers and their sizes is made for each model empirically: Several values have been tried to identify some that seem to consistently perform better than others. Other parameters, such as the number of epochs of training, the learning rate, or the batch size, are established by following the same approach. All the parameters used for training are recorded in our code repository on github. The results below were obtained using relatively small models on top of the embeddings used, the largest (Schema.org/Llama2) containing three hidden layers of sizes 100 each, and the smallest (DBpedia/RoBERTA) containing only one layer of size 15. 

All models were trained using the PyTorch Library with the Adam optimiser and the cross-entropy loss function applied to the 20-sized vector in output of the sigmoid layer, against the binary vector representing the actual relations between the input pair of entities.

\subsection{Learning individual ontologies}
\label{sec:res_ontos}

Table~\ref{tab:results} provides the results in terms of overall precision, recall, and F-score for each of the 20 trained models. To clarify here, those measures are considered on a per-relation basis, that is, if a relation exists in the ontology between a pair of entities and the model produces a number over 0.5 for the dimension of the output vector corresponding to that relation, then a true positive will be counted. If, however, the model outputs a number below the threshold of 0.5 for that relation, then a false negative is counted (similarly for true negatives and false positives).

\begin{table}[h]
    \centering
        \caption{Precision (P), recall (R) and F-score (F) (in \%) on validation sets for each of the ontologies and each of the language models used. Numbers in bold represent the best result for the corresponding measure for the ontology on that line.}
    \label{tab:results}
\begin{tabular}{l|c|rrrrr}
 &  & ~~~~~Bert~~ 
    & ~~~~RoBERTA 
    & ~~~~~GPT-2   
    & ~~~~~~Llama2      \\ \hline
gUFO         & ~~P~~ & ~69.03~          & ~62.96~~~  & ~70.95~          & ~\textbf{77.52}~  \\
             & R     & ~\textbf{87.73}~ & ~82.64~~~  & ~85.27~          & ~85.99~           \\
             & F     & ~77.27~          & ~71.47~~~  & ~77.45~          & ~\textbf{81.53}~  \\ \hline
DUL          & P     & ~84.55~          & ~84.14~~~  & ~86.40~          & ~\textbf{86.45}~  \\
             & R     & ~89.38~          & ~89.68~~~  & ~90.34~          & ~\textbf{91.00}~  \\
             & F     & ~86.90~          & ~86.83~~~  & ~88.33~          & ~\textbf{88.67}~  \\ \hline
OpenvVocab~~ & P     & ~36.21~          & ~36.16~~~  & ~36.76~          & ~\textbf{38.91}~  \\
             & R     & ~73.04~          & ~70.43~~~  & ~\textbf{80.17}~ & ~\textbf{80.17}~  \\
             & F     & ~48.41~          & ~47.79~~~  & ~50.41~          & ~\textbf{52.39}~  \\ \hline
Schema.org   & P     & ~57.06~          & ~58.35~~~  & ~56.36~          & ~\textbf{66.87}~  \\
             & R     & ~82.29~          & ~81.70~~~  & ~78.47~          & ~\textbf{84.93}~  \\
             & F     & ~67.39~          & ~68.08~~~  & ~65.60~          & ~\textbf{74.83}~  \\ \hline
DBpedia      & P     & ~42.86~          & ~37.96~~~  & ~36.57~          & ~\textbf{44.98}~  \\ 
             & R     & ~\textbf{78.41}~ & ~72.41~~~  & ~72.11~          & ~72.41~           \\
             & F     & ~55.43~          & ~49.81~~~  & ~48.53~          & ~\textbf{55.49}~  
\end{tabular}
\end{table}

\paragraph{\normalfont \textbf{Pre-trained language models capture ontological relations.}}
The first conclusion which can be drawn is that the results confirm that, to an extent, the tested pre-trained language models include sufficient information in their representation of texts to be able to recognise ontological relations between classes and properties, the best results obtained being Llama2 on DUL for an F-score of more than 88\%. This is an interesting result in itself, as it shows that, even without much effort in training, using only a few, small additional layers on top of the produced embeddings, a fairly accurate reconstruction of a significant part of some of the ontologies can be achieved. 

\paragraph{\normalfont \textbf{Llama2 performs significantly better than other language models.}}
Another straightforward conclusion from Table~\ref{tab:results} is that Llama2 performs significantly better than other language models on this task, for all measures. This is not surprising considering that even the small version of Llama2 we used is much larger than the other models. This could indicate that using a larger version of Llama2 or other larger models could potentially lead to better results, although it is obvious by observing the progression of the F-score from BERT to Llama2 that the performance is unlikely to increase linearly with the size of the model. Also, it can be seen that even if it is of the same size as BERT (and supposedly an improvement over it), RoBERTA is often the worst performing model, and that GPT-2, which can be considered as being of medium size, does not always obtain better results than BERT. The other reasons that can explain the better performance of Llama2 are the larger/better quality corpus used to train it and the larger size of the embedding vectors extracted from it, which can therefore encapsulate more information about the meaning of the words they represent. 

\paragraph{\normalfont \textbf{All ontologies are not equally predictable.}}
 A third conclusion that can be easily derived from the results obtained is that not all ontologies are equally predictable. Here, however, the potentially obvious explanation that this relates to the difference in the sizes of the training sets does not hold. The largest of the ontologies, even after filtering out over-represented properties, is DBpedia, which is close to achieving the worst results. The best results are obtained from gUFO and DUL which are, compared to the others, relatively small. Without overinterpreting the description of each ontology, a reasonable explanation could relate the performance of the prediction model to the quality of the ontology. DUL and gUFO being foundational ontologies aiming to provide a precise semantic framework to connect domain ontologies, they have been carefully, manually designed. The third best-performing ontology, schema.org, while built through community initiatives, is subject to validations to ensure that it fits its purpose. OpenVocab, on the other hand, is built through individual contributions without, it appears, much central validation, and DBpedia is first extracted from textual resources in Wikipedia, which are themselves often incomplete and possibly incorrect. It is worth mentioning, in addition, that due to our process deriving inferred relations from the stated ones, small errors in some ontologies might lead to many derived erroneous relations being included in our datasets. This is especially true for the DBpedia ontology, where a combination of an often approximate use of the domain and range relations with a large subclass hierarchy can lead to large numbers of invalid inferred domain and range relations being derived. 

\paragraph{\normalfont \textbf{Recall is greater than precision.}}
Finally, it can also be noticed that, in all cases, recall is greater than precision in the results obtained. This means that the produced models will more often generate a relation between two entities even if one should not be there (according to the original ontology), than miss predicting a relation that actually exists. There can be two explanations for this. First, ontologies might be incomplete, meaning that they might not express all the relations between all entities, and this incompleteness might not be consistent across the ontology. This could lead the model, relying on the knowledge encapsulated in the language model, to generate a relation that should be there but that the original ontology had simply missed. Second, this could partially be an artefact of only including in the datasets pairs of entities that are actually related, leaving out the many pairs between which there is no relation. This could bias the model towards generating more relations than actually exist. 

\subsection{Cross testing ontology models}
\label{sec:cross}

\begin{figure}[h]
    \centering
    \includegraphics[width=\linewidth]{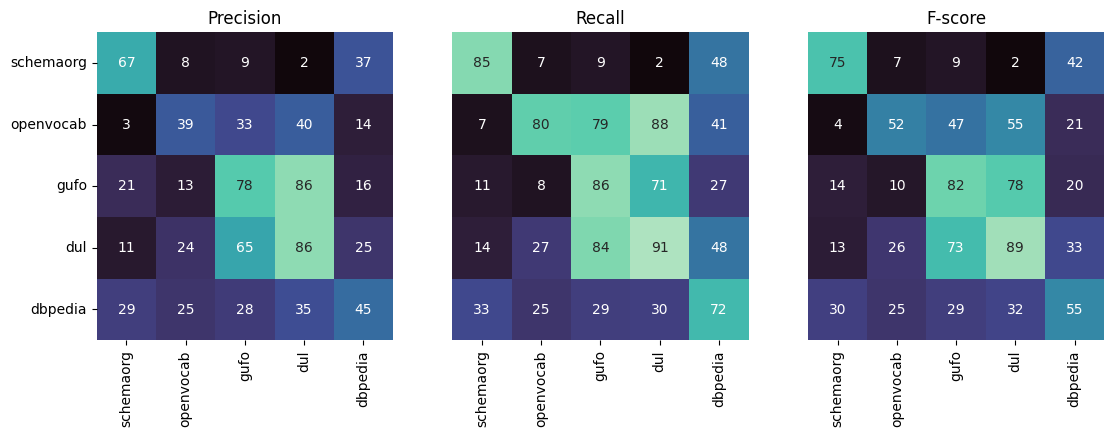}
    \caption{Precision, recall, and F-score (in \%) of testing the Llama2-based models for each ontology (lines) on the validation sets of each of the ontologies (columns).}
    \label{fig:cross-validation}
\end{figure}

In Figure~\ref{fig:cross-validation}, we present the results, in terms of precision, recall, and F-score, of cross-validating models built on the training sets of each of the ontologies, on the validation sets of each of the ontologies. Here, we rely on the five models built using Llama2, as the best performing language model for our task. The diagonals in the three matrices of Figure~\ref{fig:cross-validation} therefore correspond to the results already presented in the last column of Table~\ref{tab:results}.

The first conclusion here is that, once again without surprise, a model trained on a part of an ontology is better able to predict another part of the same ontology, rather than a part of another ontology. Beyond this obvious statement, however, we can also see that the models based on the two upper-level ontologies are not only the ones obtaining the best results on their ontology, they also generalise fairly well to predicting each other. A more surprising result is that, however, even though it generally reaches very low performances, the model trained on OpenVocab is not much worse at predicting gUFO and DUL than it is at predicting OpenVocab itself. A possible explanation for this is that the low quality of OpenVocab might not be as much the incorrectness of the relations it expresses as its incompleteness. Finally, another interesting aspect of these results is the observation that even though it achieves good performance on schema.org itself, the model trained on this ontology performs extremely badly when predicting other ontologies with the exception of DBpedia. This is likely due to the lack of domain and range relations in schema.org, which are, however, frequent in the others. It also shows that even if they are of different qualities, schema.org and DBpedia might follow similar approaches or principles for knowledge modelling that are significantly different from the ones followed by the other three ontologies. 

\subsection{Building and testing a global model}
\label{sec:combined}

As a last experiment, we trained a ``global'' model on a combined training set from the five ontologies and tested it using the five validation sets, again relying here on the Llama2 language model. The goal is to compare the performance of such a general model, trained on a larger and more diverse set of relations, with the results obtained above for more specific models, trained on smaller amounts of data. The results, in terms of precision, recall, and F-score, on the combined validation set and on each validation set individually, are presented in Table~\ref{tab:combined}. 

%% table
\begin{table}[h]
    \centering
    \caption{Precision (P), recall (R) and F-score (F) of the model trained on the combined training set, globally and per validation set  (in \%).}
    \label{tab:combined}

\begin{tabular}{l|rrrrrr}
\toprule
 & Global
 & ~~gUFO
 & ~~~DUL 
 & ~~Openvocab
 & ~~Schema.org
 & ~~DBpedia \\
\midrule
P~ & 58.17~ & 72.10~ & 86.67 & 28.40~~~~ & 58.15~~~~ & 49.27~~ \\
R~ & 72.03~ & 87.17~ & 89.64 & 19.83~~~~ & 72.60~~~~ & 65.30~~ \\
F~ & 64.36~ & 78.92~ & 88.13 & 23.35~~~~ & 64.58~~~~ & 56.16~~ \\
\bottomrule
\end{tabular}

\end{table}

As can be seen, increasing in this way the size of the training set did not lead, as could have been expected, to major improvements (the global validation measures appear close to the average scores for the previous ontology-specific models). However, it does highlight the importance of selecting the ontologies used for training on the basis of quality rather than quantity. Indeed, results for DBpedia, for example, remained similar to those obtained before, which would be expected since it is the greatest contributor to the combined training set. However, results for the two upper-level ontologies also remained similar to those obtained with their specific models, and relatively high, while they are both of smaller sizes. The results for Schema.org, on the other hand, dropped several points, while it represents the second biggest training set. This shows that including in the training set ontologies that do not follow the same design principles or are differently aligned in their textual representation and conceptual representation with the language model might lead to difficulties for such a simple prediction model as the one we used here. OpenVocab is an even clearer example of this, showing disastrous results, probably because it is of significantly different quality from the others and of small size. 

Finally, in Table~\ref{tab:perrel} the precision, recall, and F-score of the combined Llama2-based model are reported for each of the considered relations individually.

\begin{table}[h]
\centering
        \caption{Precision (P), recall (R) and F-score (F) for specific relations (in \%). Relations for which results were too rare to give meaningful measures were removed.}
    \label{tab:perrel}
{\small
\setlength\tabcolsep{1.5pt}
\begin{tabular}{c|rrrrrrrrrrrrr}
\toprule
 & SpC~  & SpP~  & HR~   & RO~ 
 & ISbC~ & ISpC~ & ISbP~ & ISpP~ 
 & IRO~  & IHR~  & IDO~  & IHD~ \\
\midrule
P~ & 20.12~ & 66.23~ & 18.66~ & 50.00~
   & 65.32~ & 73.55~ & 59.26~ & 64.71~ 
   & 66.89~ & 28.29~ & 52.43~ & 35.21~ \\
R~ & 48.95 & 40.0 & 30.12 & 9.09 & 82.39 & 86.73 & 33.22 & 38.19 & 82.23 & 41.95 & 95.31 & 48.45 \\
F~ & 28.52 & 49.88 & 23.04 & 15.38 & 72.87 & 79.6 & 42.57 & 48.03 & 73.77 & 33.8 & 67.65 & 40.78 \\
\bottomrule
\end{tabular}
}
\end{table}

We can observe from this table that, as expected, not all relations are predicted equally. First, several relations, as can be seen in Table~\ref{tab:relations}, are simply too rare in the training set to be properly learnt. Also, it appears clearly that direct, stated relations (e.g. SpC) are harder to predict than their inferred, indirect counterparts (e.g. ISpC). This can be explained partly by the fact that there are many more inferred relations in the training set than direct ones, but also that learning to recognise that a relation not only exists but is direct in the considered ontology is naturally harder. Finally, we can see a surprising unbalance between some relations that are semantically related. For example, super-classes and super-properties seem to be better recognised than sub-classes and sub-properties. Similarly, the model appears to be better at predicting that a class is the domain or range of a property than that a property has a class for domain or range. Once again, the lesson here is that future models for predicting ontological relations from multiple ontologies would potentially require not only a more complex neural architecture, but also a more diverse and balanced training set from which to learn.

\section{Discussion: possible applications}
\label{sec:discussion}
The results presented in the previous section show a promising new way in which semantic web tools and applications could effectively exploit web knowledge, as captured by the kind of models created here. Before the emergence of large language models, great potential was attributed to semantic web search engines~\cite{daquin_semantic_2011} for their ability to centralise ontological knowledge on the Web and therefore act as a hub for applications aiming to exploit such knowledge. However, those search engines have for the most part disappeared, as they were complex systems that were difficult to maintain. They also suffered from many limitations, in particular with respect to response time, but also to the inherent limits of available web ontologies. The results presented here show that, while they are different in nature, models for predicting ontological relations built on a large and diverse set of ontologies, being proper large knowledge models, could partially fulfil this potential by capturing in an efficient form the core content of those ontologies, with the added ability to, at least partly, extrapolate from them. The most obvious of applications for such models would therefore be ontology matching. Indeed, while the idea of using a language model for this task, finding relations between entities of different ontologies, is not new (see, for example,~\cite{hertling_olala_2023}), integrating an ontology relation prediction model would appear to be a straightforward way to integrate large amounts of web knowledge into the process, in a form that could also be easily combined with other approaches. Similarly, such models could be used to support the construction of ontologies or their knowledge-based evolution (in a way similar to the one presented in~\cite{aroyo_ontology_2009}) by being able to suggest where to place a new class or a new property in relation to existing ones. This could, in addition, use the particular form of the model's output, a number for each possible relation that is closer to 1.0 when it is more likely that the relation exists, to offer multiple relations ranked by their likeliness. It is not hard to see how a component building on the models presented here (and on any future ones of the same nature) could be integrated into many ontology engineering tools and applications. Beyond the ontology engineering field, however, we can also imagine how large knowledge models, learnt from web ontologies, could be integrated in other applications of language models, enabling them to benefit not only from knowledge expressed in texts, but also from knowledge expressed formally in those ontologies. 

\section{Conclusion}
\label{sec:conclusion}

In this paper, we report on experiments to build neural models to predict ontological relations (direct or inferred) between classes and properties from word embeddings. We showed that even very simple models built on top of such embeddings for the textual representation of those entities obtained promising results. We also showed that even if the results were often similar, the larger Llama2 model was consistently better as a source of embeddings in this task than other smaller models. We also discussed how the results were largely dependent on the quality of the ontology(ies) on which the model was trained, with carefully designed, upper-level ontologies leading to excellent results where unvalidated, community built ontologies led to disappointing model performances. 

Based on the promising results obtained, we discussed possible applications of models that could, improving upon those presented here, encapsulate and extrapolate from the knowledge contained in web ontologies. There are many avenues for such improvements that can be enabled by the resources created and made available for this paper. First, as mentioned above, building ontology relation prediction models on a larger and better selected corpus of ontologies could lead to higher quality predictions. Collecting such a large corpus and setting up mechanisms by which ontologies can be selected are therefore part of our ongoing work. Also, with a larger and more diverse corpus, it is to be expected that other, potentially more complex models might be necessary, and there remain many variations of the parameters, model architectures, training regimes, etc. that could be tested on this task. Similarly, different approaches to combining the models learnt from specific ontologies could be considered, from simple ensemble methods to more advanced mixture of experts approaches~\cite{masoudnia_mixture_2014}. Finally, an obvious way in which improvements in results could be obtained is by including an element of fine-tuning of the language models used in training for the task of predicting ontological relations, although such an approach would have higher requirements with respect to available computing resources. 

\section*{Supplementary material}
\begin{sloppypar}
At \texttt{\url{https://figshare.com/articles/dataset/Data_and_models_for_Ontological_relations_from_word_embeddings_/25601010?file=45645084}} is the FigShare data repository that includes the built models and the measures obtained from their validation. It also includes the generated datasets used as input to the training and validation steps and the intermediary structures built as part of constituting those datasets (indexes and matrices). The ontologies themselves are not included, but a metadata file indicates from where they were downloaded, at what time.
\end{sloppypar}

% At \texttt{\url{https://anonymous.4open.science/r/LKM-B71B}} the code and configuration files to build the datasets, train the models, and validate them are available in a github repository.

 \bibliographystyle{plain}
 \bibliography{zoteroClean}

\end{document}